\documentclass{article}

\usepackage{PRIMEarxiv}

\usepackage[utf8]{inputenc} 
\usepackage[T1]{fontenc}    
\usepackage{hyperref}       
\usepackage{url}            
\usepackage{booktabs}       
\usepackage{amsfonts}       
\usepackage{nicefrac}       
\usepackage{microtype}      
\usepackage{lipsum}
\usepackage{fancyhdr}       
\usepackage{graphicx}       
\graphicspath{{media/}}     

\usepackage[numbers]{natbib}
\usepackage{multirow}
\usepackage{caption}
\usepackage{subcaption}
\usepackage{epsfig}
\newcommand{\tabincell}[2]{\begin{tabular}{@{}#1@{}}#2\end{tabular}}

\title{ATLANTIS: A Benchmark for Semantic Segmentation of Waterbody Images
}

\author{
  Mohammad H. Erfani \\
  Department of Civil and Environmental Engineering \\
  University of South Carolina \\
  Columbia, SC\\
  \texttt{serfani@email.sc.edu} \\
   \And
  Zhenyao Wu, Xinyi Wu \\
  Department of Computer Science and Engineering \\
  University of South Carolina \\
  Columbia, SC\\
  \texttt{\{zhenyao, xinyiw\}@email.sc.edu} \\
    \And
  Song Wang \\
  Department of Computer Science and Engineering \\
  University of South Carolina \\
  Columbia, SC\\
  \texttt{songwang@cec.sc.edu} \\
    \And
  Erfan Goharian \\
  Department of Civil and Environmental Engineering \\
  University of South Carolina \\
  Columbia, SC\\
  \texttt{goharian@cec.sc.edu} \\
}

\begin{document}
\maketitle

\begin{figure}[htbp]
    \centering
    \makebox[\linewidth][c]{%
    \includegraphics[width=1.0\columnwidth]{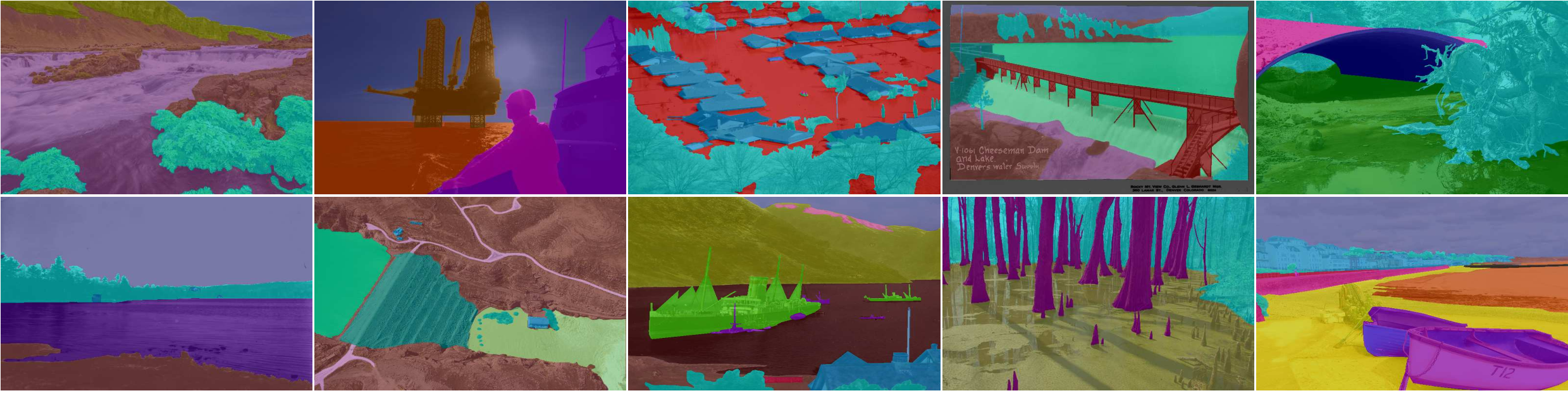}}
    \caption{ATLANTIS: ArTificiaL And Natural waTer-bodIes dataSet.}
    \label{fig:atlantis}
\end{figure}

\begin{abstract}
    Vision-based semantic segmentation of waterbodies and nearby related objects provides important information for managing water resources and handling flooding emergency. However, the lack of large-scale labeled training and testing datasets for water-related categories prevents researchers from studying water-related issues in the computer vision field. To tackle this problem, we present ATLANTIS, a new benchmark for semantic segmentation of waterbodies and related objects. ATLANTIS consists of 5,195 images of waterbodies, as well as high quality pixel-level manual annotations of 56 classes of objects, including 17 classes of man-made objects, 18 classes of natural objects and 21 general classes. 
    We analyze ATLANTIS in detail and evaluate several state-of-the-art semantic segmentation networks on our benchmark. In addition, a novel deep neural network, AQUANet, is developed for waterbody semantic segmentation by processing the aquatic and non-aquatic regions in two different paths. AQUANet also incorporates low-level feature modulation and cross-path modulation for enhancing feature representation. Experimental results show that the proposed AQUANet outperforms other state-of-the-art semantic segmentation networks on ATLANTIS. We claim that ATLANTIS is the largest waterbody image dataset for semantic segmentation providing a wide range of water and water-related classes and it will benefit researchers of both computer vision and water resources engineering.
\end{abstract}


\section{Introduction}
Every year, floods claim tens of billions of US dollars losses and thousands of lives globally~\cite{hirabayashi2013global}. Accurate detection, measurement, and track of the waterbodies can help both the public and decision-makers to take appropriate actions to minimize the risk and losses~\cite{huang2018near, gebrehiwot2019deep}. With the popularity of smart phones and drones, various data at flooding sites can be collected rapidly and continuously to provide more useful and heterogeneous information source~\cite{eltner2021using, eltner2018automatic, cervone2016using, schnebele2013improving}, compared to the conventional gauge sensing and remote sensing~\cite{eltner2021using, lo2015visual}. As a fundamental step to leverage such collected images for modeling and decision-making, we need first to conduct a refined semantic segmentation of included waterbodies and related objects in such scenes, which we focus on in this paper.

With the advancement of deep neural networks, semantic segmentation has achieved a great success in recent years on various kinds of images, such as natural images~\cite{lin2014microsoft}, street images~\cite{cordts2016Cityscapes, yu2020bdd100k}, and medical images~\cite{bernal2017comparative, jha2020kvasir}. However, waterbody images pose many new unique challenges for semantic segmentation. 
In some forms, water preserves the intrinsic properties such as reflection, transparency, shapeless and colorless visual features; which in turn, brings difficulties to semantic segmentation of water and related objects. Moreover, in some other forms, these properties can be affected by illumination sources from surroundings, turbidity, and turbulence. Different-labeled waterbodies, such as river and canal, or lake and reservoir, often have similar visual characteristics that make task of semantic segmentation even harder.
As shown in our later experiments, these unique challenges may significantly affect the performance of the existing semantic segmentation networks.

Meanwhile, deep-learning based approaches always require large-scale training data with necessary ground-truth annotations. Lack of such a public dataset for waterbody segmentation significantly impedes the research on this problem. The collection and annotation of such a dataset can be very laborious and time-consuming to cover a wide range of waterbodies and related objects. There is no specific repository providing relevant images. In addition, team members and annotators are required to have prior knowledge on water resources engineering to be capable of selecting and precisely annotating the images.

In this paper, we present a new benchmark, ATLANTIS (ArTificiaL And Natural waTer-bodIes dataSet). For the first time, this dataset has covered a wide range of natural and man-made (artificial) waterbodies such as sea, lake, river, canal, reservoir, and dam. ATLANTIS includes 5,195 pixel-wise annotated images split to 3,364 training, 535 validation and 1,296 testing images. As shown in the Table~\ref{tab:labels_list}, in addition to 35 waterbody and water-related objects, ATLANTIS also covers 21 general labels. Moreover, we construct ATLANTIS Texture (ATeX) dataset, which consists of 12,503 patches for the water-bodies texture classification, sampled from 15 kinds of waterbodies in ATLANTIS.

\begin{table}[htbp]
    \centering 
    \caption{List of the ATLANTIS labels.}
    \label{tab:labels_list}
    \begin{tabular}{ll} 
    \toprule
    Artificial & \tabincell{l}{breakwater; bridge; canal; culvert; dam; ditch; levee; lighthouse; pipeline; \\ 
                               pier; offshore platform; reservoir; ship; spillway; swimming pool; \\ 
                               water tower; water well.} \\
    \midrule
    Natural &   \tabincell{l}{cliff; cypress tree; fjord; flood; glaciers; hot spring; lake; mangrove; marsh; \\
                              puddle; rapids; river; river delta; sea; shoreline; snow; waterfall; wetland.}\\ 
    \midrule
    General & \tabincell{l}{road; sidewalk; building; wall; fence; pole; traffic sign; vegetation; terrain; \\ 
                            sky; train; person; car; bus; truck; bicycle; parking meter; motorcycle; \\
                            fire hydrant; boat; umbrella.}\\ 
    \bottomrule
    \end{tabular}
\end{table}

In order to tackle the inherent challenges in the segmentation of waterbodies, AQUANet is developed which takes an advantage of two different paths to process the aquatic and non-aquatic regions, separately. Each path includes low-level feature and cross-path modulation, to adjust features for better representation. The results show that the proposed AQUANet outperforms other ten state-of-the-art semantic-segmentation networks on ATLANTIS, and the ablation studies justify the effectiveness of the components of the proposed AQUANet.

\section{Related Work}

\subsection{Semantic segmentation dataset}
Large-scale annotated datasets, such as COCO~\cite{lin2014microsoft}, PASCAL Context~\cite{mottaghi2014role}, ADE20K~\cite{zhou2019semantic}, and more recently Mapillary Vistas Dataset~\cite{neuhold2017mapillary} and BDD100K~\cite{yu2020bdd100k}, make it possible for researchers to develop deep-learning based models for real-world applications. Considering the most related dataset to ATLANTIS,~\citeauthor{gebrehiwot2019deep}~\cite{gebrehiwot2019deep} collected a small number of top-view waterbody dataset (100 images) using Unmanned Aerial Vehicles (UAVs) which contains only four categories (i.e., water, building, vegetation and road). In addition,~\citeauthor{sazara2019detecting}~\cite{sazara2019detecting} introduced a larger dataset (253 images) which just focuses on flood region segmentation. More recently,~\citeauthor{sarp2020detecting}~\cite{sarp2020detecting} provided a dataset that consists of 441 annotated roadway flood images. However, these datasets have limitations in either the number of annotated images, or the categories they covered, and none of those considers more complex classes of waterbodies such as sea, lake and waterfall. Therefore, we develop a new dataset, ATLANTIS, as the first large-scale annotated dataset to provide a wide-range of waterbodies and water-related objects.

\subsection{Semantic segmentation network}
All of existing semantic segmentation approaches share the same goal to classify each pixel of a given image but differ in the network design, including low-resolution representations learning~\cite{long2015fully,chen2017deeplab}, high-resolution representations recovering~\cite{badrinarayanan2015segnet,noh2015learning,lin2017refinenet}, contextual aggregation schemes~\cite{YuanW18,zhao2017pyramid,yuan2019object}, feature fusion and refinement strategy~\cite{lin2017refinenet,huang2019ccnet,li2019expectation,zhu2019asymmetric,fu2019dual}. 
Typically, method designs are dependent on their respective datasets and all the mentioned networks are developed by training on benchmark datasets such as Cityscapes~\cite{cordts2016Cityscapes}, COCO~\cite{lin2014microsoft} and VOC~\cite{everingham2010pascal} where the inter-class boundary is clear even for the within-group categories (e.g., car and truck). As mentioned above, waterbody images pose new challenges to semantic segmentation. Previous works on waterbody segmentation mainly use satellite imagery~\cite{munoz2021local, li2019multiscale, duan2019multiscale}. In this work, we focus on natural waterbody images terrestrially captured by various cameras, and design AQUANet, a new two-path semantic segmentation network, by including an aquatic branch explicitly for waterbody classes.

\begin{figure}[htbp]
     \centering
     \begin{subfigure}[b]{0.9\textwidth}
         \centering
         \includegraphics[width=\textwidth]{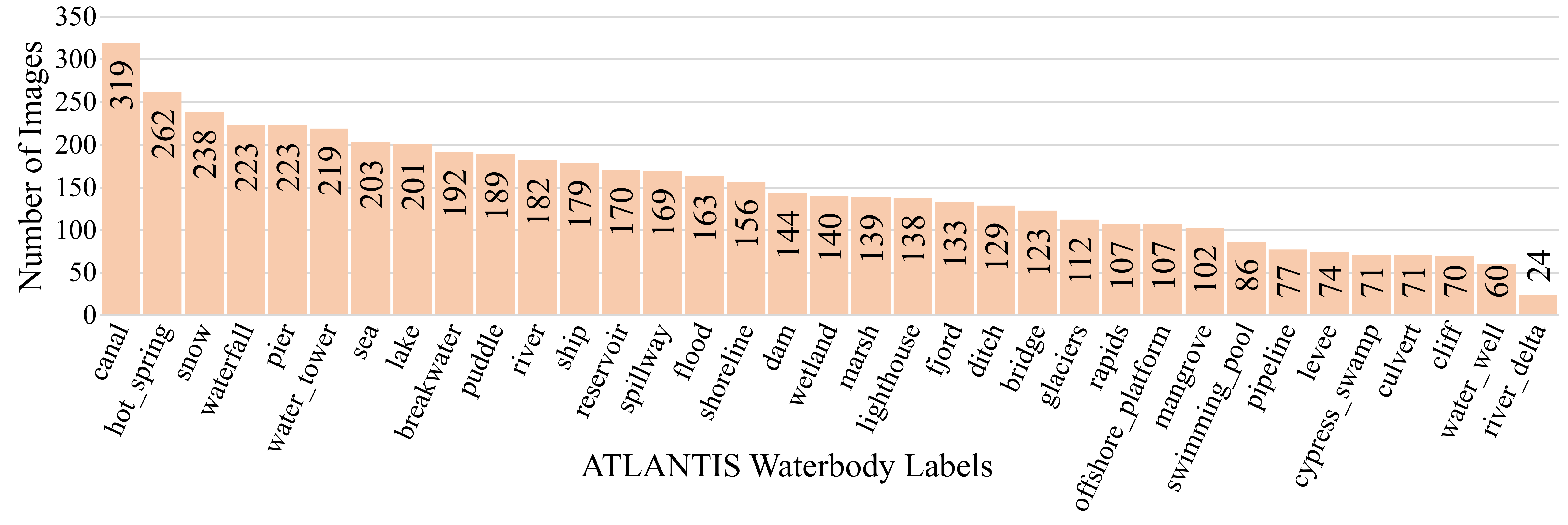}
         \vspace{-0.5em}
         \caption{}
         \label{sfig:freq-imgs}
     \end{subfigure}
     \begin{subfigure}[b]{0.9\textwidth}
         \centering
         \includegraphics[width=\textwidth]{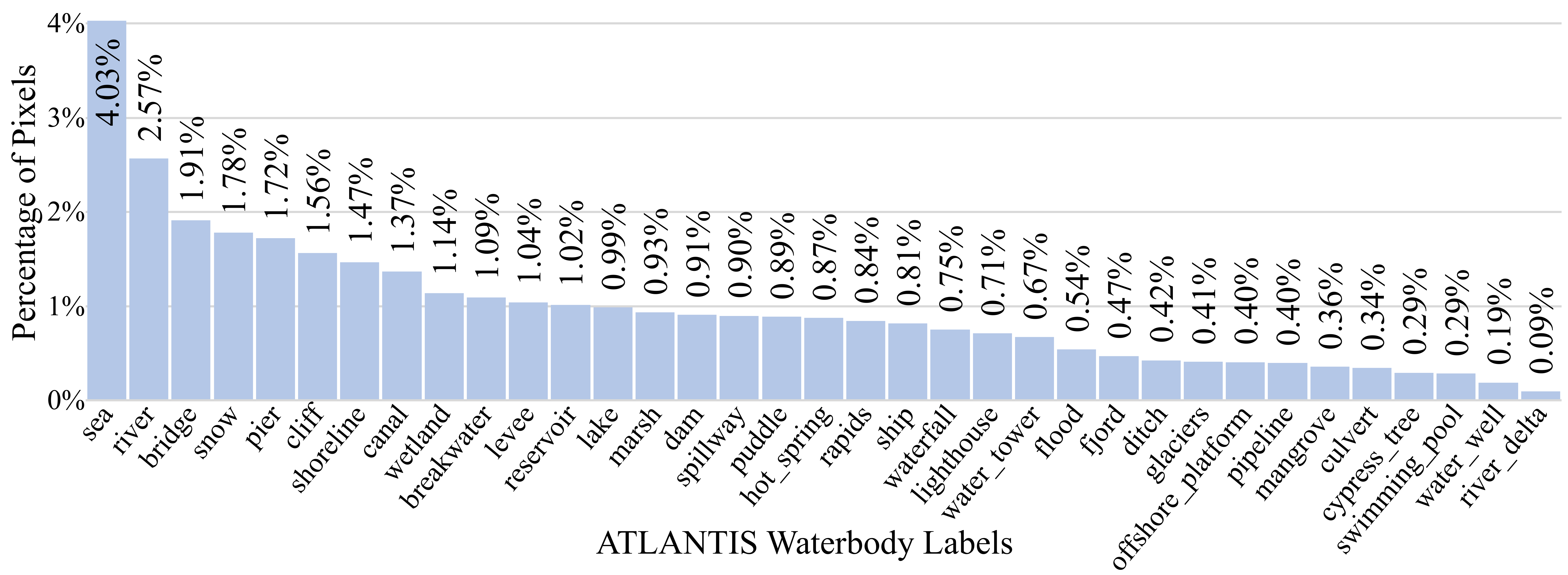}
         \vspace{-0.5em}
         \caption{}
         \label{sfig:percentage-pixels}
     \end{subfigure}
     \vspace{-0.5em}
        \caption{(a) The frequency distribution of images for different waterbodies in ATLANTIS (b) The Percentage of pixels for waterbody labels.}
        \label{fig:freq-imgs-pixels}
\end{figure}

\section{ATLANTIS Dataset} \label{sec:dataset}
The ATLANTIS dataset is designed and developed with the goal of capturing a wide-range of water-related objects, either those exist in natural environment or the infrastructure and man-made (artificial) water systems. In this dataset, labels were first selected based on the most frequent objects, used in water-related studies or can be found in real-world scenes. Aside from the background objects, total of 56 labels, including 17 artificial, 18 natural waterbodies, and 21 general labels, are selected (Table~\ref{tab:labels_list}). These general labels are considered for providing contextual information that most likely can be found in water-related scenes. After finalizing the selection of waterbody labels, a comprehensive investigation on each individual label was performed by annotators to make sure all the labels are vivid examples of those objects in real-world. Moreover, sometimes some of the water-related labels, e.g., levee, embankment, and floodbank, have been used interchangeably in water resources field; thus, those labels are either merged into a unique group or are removed from the dataset to prevent an individual object receives different labels.

In order to gather a corpus of images, we have used Flickr API to query and to collect ``medium-sized'' unique images for each label based on eight commonly used ``Creative Commons'', ``No Known Copyright Restrictions'' and ``United States Government Work" licenses. Downloaded images were then filtered by a two-stage hierarchical procedure. In the first stage, each annotator was assigned to review a specific list of labels and remove irrelevant images based on that specific list of labels. In the second stage, several meetings were held between the entire annotation team and the project coordinator to finalize the images which appropriately represent each of 56 labels.

This sieving procedure has been applied four times in order to meet the limit and to reach the current number of images. The percentage of image acceptance rate for the third and fourth phases are 14.41$\%$ and 5.06$\%$, respectively. It means if we want to add 1000 more images to the dataset, we should process at least 20,000 images. Finally, images were annotated by annotators who have solid water resources engineering background as well as experience working with the CVAT~\cite{boris_sekachev_2020_4009388}, which is a free, open source, and web-based image/video annotation tool.

\subsection{Dataset statistics} \label{subsec:dataset-statistics}
\begin{figure}[htbp]
    \centering
    \includegraphics[width=0.7\columnwidth]{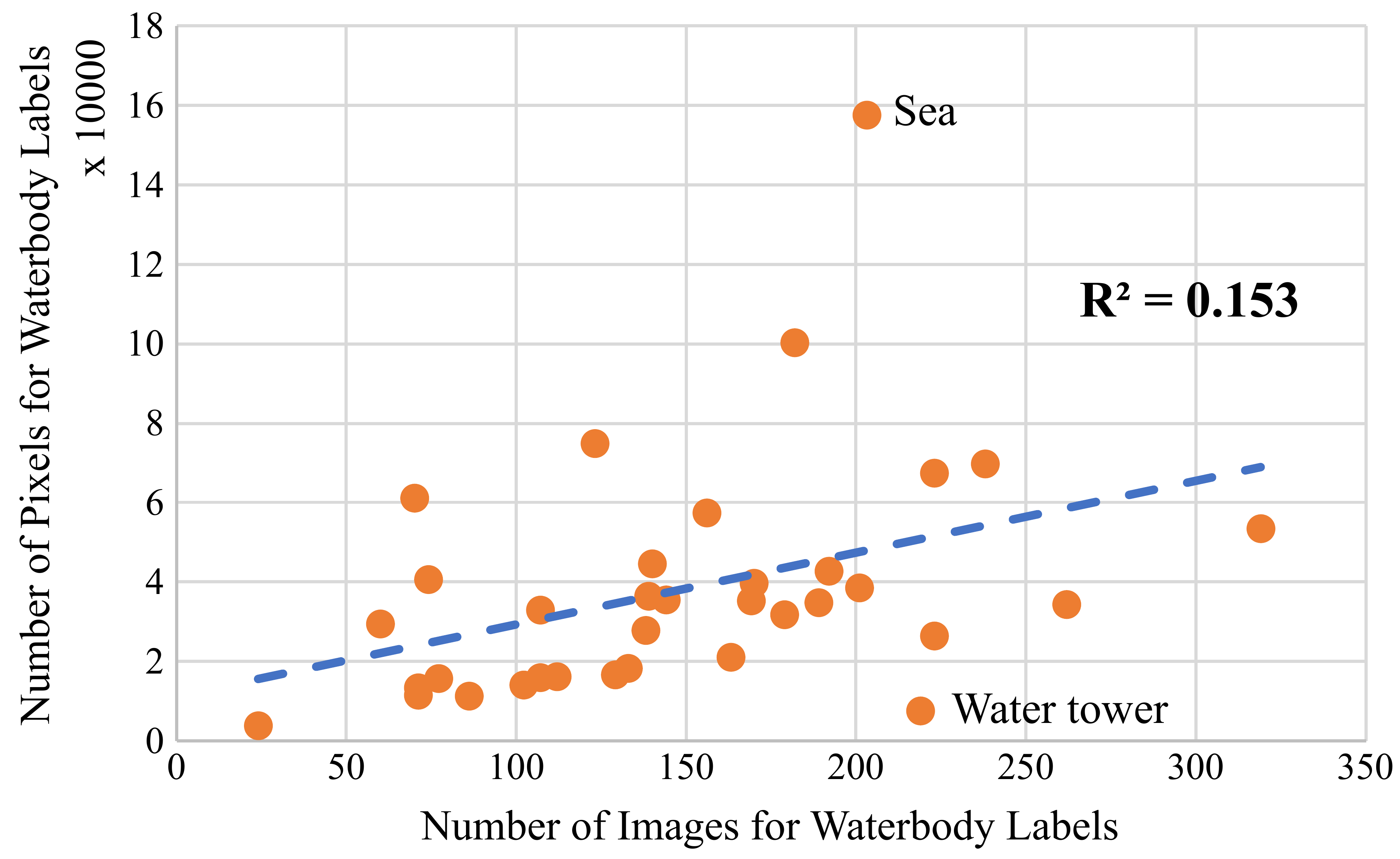}
    \caption{Correlation between number of images for each label and the corresponding pixels}
    \label{fig:freq-vs-pixels}
\end{figure}

Figure~\ref{fig:freq-imgs-pixels} shows the frequency distribution of the number of images and the percentage of pixels for waterbody labels. Such a long-tailed distribution is common for semantic segmentation datasets~\cite{lin2014microsoft, zhou2019semantic} even if the number of images that contain specific label are pre-controlled. Such frequency distribution for pixels would be inevitable for objects existing in real-world. Taking ``water tower'' as an example, despite having 219 images, the percentage of pixels are less than many other labels in the dataset. Figure~\ref{fig:freq-vs-pixels} shows the positive but weak correlation between number of images for each label and the corresponding pixels. In total, only $4.89\%$ of pixels are unlabeled, and $34.17\%$ and $60.94\%$ of pixels belong to waterbodies (natural and man-made) and general labels, respectively. As it is evident, the main proportion of pixels belongs to general labels. This clearly shows the importance of general labels for better scene understanding~\cite{caesar2018coco} and accurate object classification in semantic segmentation network.

\subsubsection{Spatial analysis} \label{subsubsec:spatial-analysis}
\begin{figure}[htbp]
    \centering
    \includegraphics[width=0.7\linewidth]{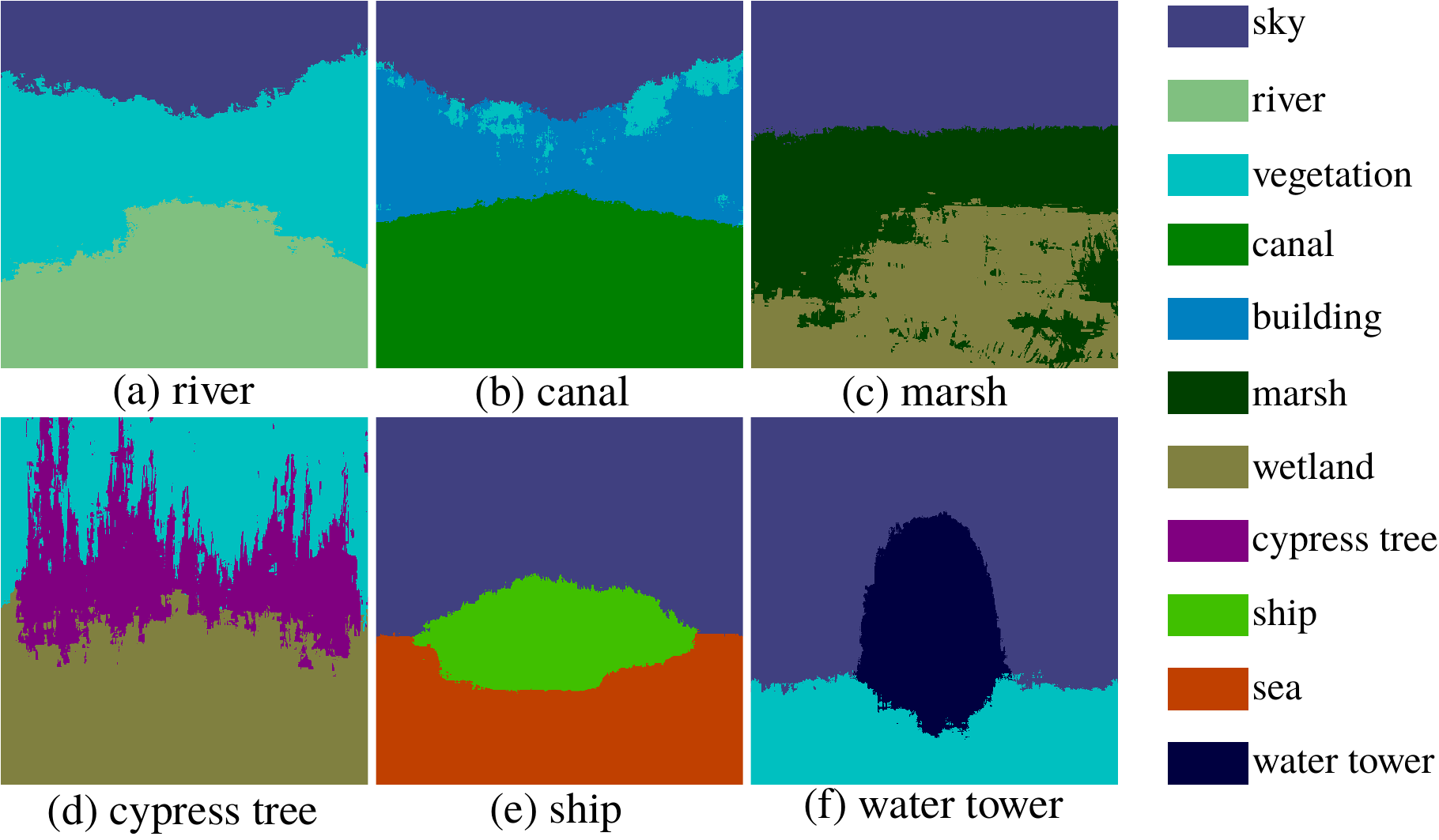}
    \caption{Spatial analysis of four different waterbody labels.}
    \label{fig:spatial-analysis}
\end{figure}

Following ADE20K dataset~\cite{zhou2019semantic}, a spatial analysis, known as ``mode of the object segmentation'' has been done on the ground truth segmentation map for each label. Specifically, considering a waterbody label $L$ with $n$ images in ATLANTIS, we resize their corresponding $n$ ground-truth segmentation maps to $512 \times 512$ pixels. We then count the most frequent label at each pixel of the map and construct a ``mode of segmentation'' map for label $L$, as shown in Figure~\ref{fig:spatial-analysis}. This map demonstrates the spatial distributions of the most frequent co-occurred labels with respect to a given waterbody label. Based on this, we equipped our proposed network with cross-path feature modulation to cope with the difficulties associated with the recognition of waterbody labels having visual similarities.

\subsection{Image annotation} \label{subsubsec:image-annotation}

\subsubsection{Annotation pipeline}
\begin{figure}[htbp]
    \centering
    \begin{subfigure}[b]{0.595\columnwidth}
        \centering
        \includegraphics[width=\textwidth]{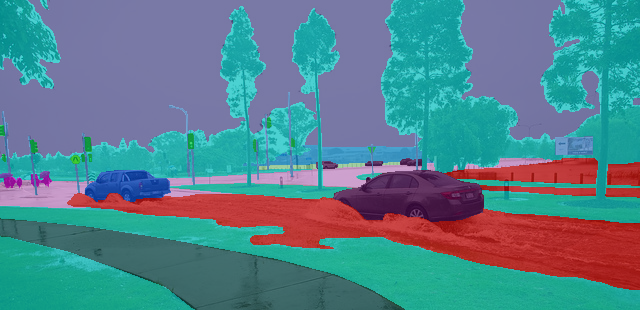}
        \label{sfig:flood1}
    \end{subfigure}
    \begin{subfigure}[b]{0.385\columnwidth}
        \centering
        \includegraphics[width=\textwidth]{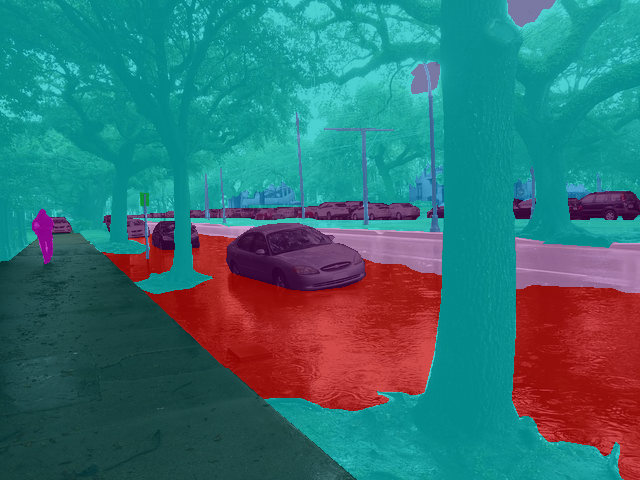}
        \label{sfig:flood2}
    \end{subfigure}
	\vspace{-1em}
    \caption{Two samples of complex flood scenes}
    \label{fig:flood-scene}
\end{figure}

ATLANTIS was annotated by six annotators having prior knowledge in the area of water resources. The goal in the annotation task is to balance speed and quality. Generally, time spent on a single image can range from 4 minutes to 25 minutes depending on the image complexity. In this project, each kind of waterbody was assigned to a specific annotator. Before annotation of a label, all the images of that label are scrutinized and discussed by a group of experts in water resources engineering. 
We can see that the annotation of complex flood scenes takes more time since such images are usually captured in urban areas and have more elaborated components as shown in Figure~\ref{fig:flood-scene}.  

\subsubsection{Consistency analysis}
While one image is annotated by one annotator for ATLANTIS, we perform additional consistency analysis across annotators and over time for an annotator. We choose 52 images from ATLANTIS, by including both images that are highly susceptible to wrong labelling and those contain objects prone to be either left unannotated or wrongly annotated. We ask three annotators to annotate them again and compare the results against the already approved ground truth in ATLANTIS. The accuracy and mIoU in terms of all 52 images (total) and the subsets of images that had been annotated by themselves before (individual) are shown in Table~\ref{tab:inconsistency_results}. We can see that an annotator can process the images that he/she annotated before with much better consistency. 

\begin{table}[h]
	\centering
	\caption{Consistency analysis for three annotators.}
	\label{tab:inconsistency_results}
	\begin{tabular}{cc|c|c|c}
		\hline
		& & Annotator 1 & Annotator 2 & Annotator 3 \\
		\hline
		\multirow{2}{*}{Total} 
		& acc & 84.00  & 79.93 & 79.00 \\
		& mIoU & 62.29 & 59.33 & 57.34  \\
		\hline
		\multirow{2}{*}{Individual} 
		& acc & 91.11 & 90.44  & 94.69  \\
		& mIoU & 72.83  & 76.14 & 75.69  \\
		\hline
	\end{tabular}
\end{table}

\subsection{ATLANTIS Texture (ATex)}
\begin{figure}[ht]
    \centering
    \includegraphics[width=0.7\linewidth]{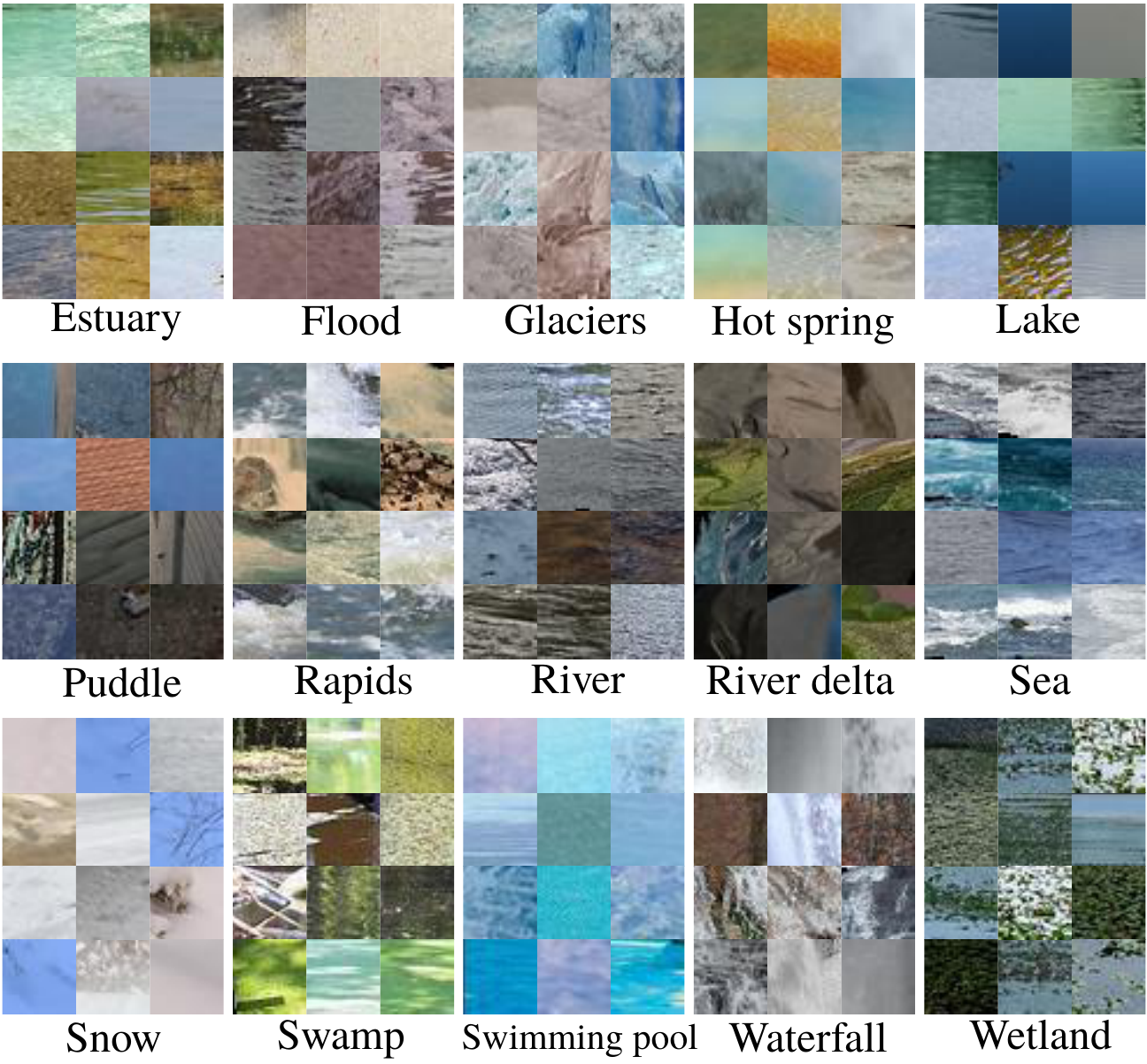}
    \caption{Samples of ATeX texture images.}
    \label{fig:atex}
\end{figure}

Waterbodies usually bear texture appearance and it is an interesting problem to study whether different kinds of waterbodies may show subtle differences associated with the texture features. For this purpose, we construct a new waterbody texture dataset, ATeX, by cropping patches from ATLANTIS and take the corresponding annotated waterbody label as the label of the patch. We set patch size to $32 \times 32$ pixels and all pixels in a cropped patch must have the same waterbody label in the original image. We also ensure there is no partial overlap between any two patches. In total we collected 12,503 patches with 15 waterbody labels: Two waterbody labels ``estuary'' and ``swamp'' are added based on the nearby tree species -- mangrove ``estuary'' and cypress for ``swamp'', while four waterbody labels ``canal'', ``ditch'', ''reservoir'' and ``fjord'' are omitted because of high visual similarities with other labels. Sample images of ATeX dataset are shown in Figure~\ref{fig:atex}. We split ATeX into 8,753 for training, 1,252 for validation and 2,498 for testing. In the later experiment, we train different models to evaluate their classification performance on ATeX images.

\section{AQUANet}
\begin{figure}[ht]
	\centering
	\includegraphics[width=1.0\linewidth]{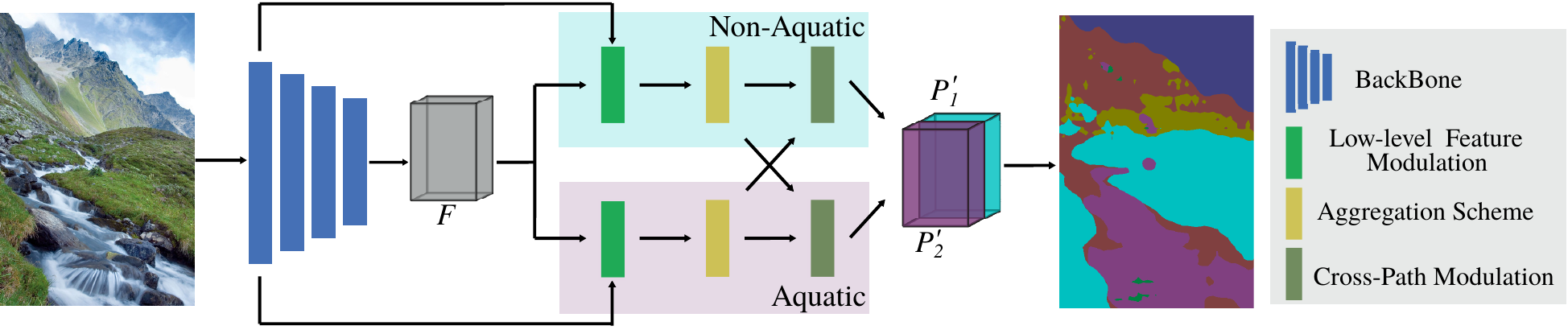}
	\caption{The network architecture of proposed AQUANet.}
	\label{fig:nework}
\end{figure}

Typically, existing semantic segmentation networks are designed based on a strong backbone (e.g. ResNet \cite{he2016deep}) to extract features from images with additional feature aggregation schemes such as ASP-OC \cite{YuanW18} and PPM \cite{zhao2017pyramid}.
Because of difficulties associated with semantic segmentation of waterbodies, we design AQUANet to segment aquatic and non-aquatic categories, separately, as shown in Figure \ref{fig:nework}. 

\subsection{Network architecture}
According to Figure \ref{fig:nework}, the input image is first fed into a ResNet-101 (pretrained on ImageNet \cite{deng2009imagenet}) to extract the feature $F$ with a size of $C \times H \times W$.
Then, the feature is sent into two separate paths for further processing. 
The aquatic path is to segment different types of waterbodies including sea, river, waterfall, wetland and etc., while the non-aquatic path is to segment other categories such as ship and bridge. 
In each path, the feature $F$ is first modulated by the low-level feature $F_l$ extracted from the third convolutional layer of ResNet-101, and then passed into ASP-OC \cite{YuanW18} to produce the probability map. 
In the last step, two cross-path modulation blocks are applied to adjust the probability maps $P_1$ and $P_2$ in parallel. Finally, the resulted probability maps are concatenated and upsampled to the size of the original image.

\subsection{Feature modulation}
\begin{figure}[htbp]
	\centering
	\includegraphics[width=0.7\linewidth]{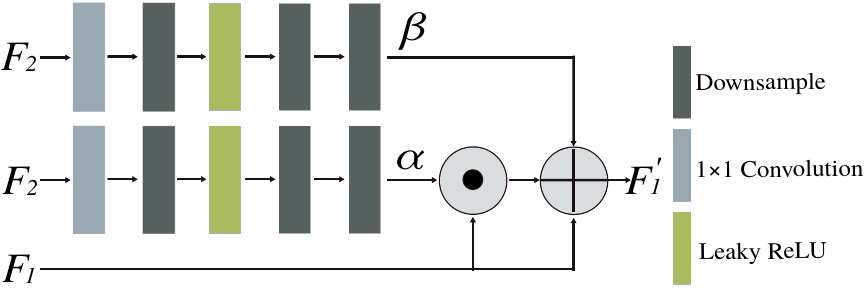}
	\caption{An illustration of the feature modulation. }
	\label{fig:modulataion}
\end{figure}

The goal of the feature modulation $\mathcal{M}$ is to adjust a feature map $F_1$ given feature map $F_2$ to represent the adjusted feature $F_1'$. It can be formulated as:
\begin{equation} 
F_1' = \mathcal{M}(F_1| F_2).
\end{equation}
To generate the modulated feature $F_1'$, the parameters $\alpha$ and $\beta$ are learned from $F_2$ via the feature modulation that consists of three downsampling layers, six $1 \times 1$ convolutional layers and two leakyReLU layers as shown in Figure \ref{fig:modulataion}. 
The learned parameters $\alpha$ and $\beta$ have the shape as the $F_1$. Then, the resulting feature $F_1'$ is constructed as follows according to~\cite{wang2018recovering, park2019SPADE}:
\begin{equation} 
F_1' = \alpha \cdot F_1 + \beta + F_1.
\end{equation}

\subsubsection{Low-level feature modulation}

To enhance the low-level texture representation of the water-bodies, we propose to use the low-level feature $F_l$ to modulate the feature $F$ and the resulted feature $F'$ is defined as:
\begin{equation} 
F' = \mathcal{M}(F | F_l).
\end{equation}
Note that different channels of the receptive field in the third convolutional layer of ResNet-101 are used to construct the low-level feature $F_l$ for the two paths.

\subsubsection{Cross-path modulation}
We also propose a cross-path modulation to aggregate the outputs of the probability maps. The adjusted probability maps resulted from this module are defined as:
\begin{equation} 
P'_1 = \mathcal{M}(P_1 | P_2),
\end{equation}
\begin{equation} 
P'_2 = \mathcal{M}(P_2 | P_1),
\end{equation}
where the $P'_1$ and $P'_2$ represent the final probability maps for the aquatic and non-aquatic objects, respectively. Note that the modulation layer is not sharing weights across the two paths.

\subsection{Loss function}

The proposed network is trained in a fully supervised fashion constrained by the widely-used cross-entropy loss function on both final prediction $P$ (main loss) and the intermediate feature produced by the fourth block of the ResNet-101 (auxiliary loss). Following \citeauthor{zhao2017pyramid}~\cite{zhao2017pyramid}, the weights of the main and the auxiliary loss are set to 1 and 0.4, respectively.

\section{Experiments}

To demonstrate the effectiveness of the proposed method for water-bodies semantic segmentation, we train AQUANet on the proposed ATLANTIS dataset.
For performance evaluation, we take the mean of class-wise intersection over union (mIoU) and the per-pixel accuracy (acc) as the main evaluation metrics.
To further evaluate the performance on the waterbodies, we calculate the mean IoU for aquatic categories (A-mIoU) and the accuracy in the aquatic region (A-acc). Aquatic categories include 17 labels showing just water content in different forms and bodies, e.g., sea, river, lake, etc.

\begin{table*}[ht]
	\centering
	\caption{The per-category results on the ATLANTIS test set by current state-of-the-art methods and our AQUANet. The best and the second best results are highlighted with \textbf{bold} font and \underline{underline}, respectively.}
	\label{tab:test}
	\footnotesize
	\setlength{\tabcolsep}{0.7mm}
	    \makebox[\linewidth][c]{%
		\begin{tabular}{l|ccccccccccccccccc|c|c|c|c}
			\toprule
			Method & \rotatebox{90}{canal}  & \rotatebox{90}{ditch} & \rotatebox{90}{fjord} & \rotatebox{90}{flood} & \rotatebox{90}{glaciers} & \rotatebox{90}{hot spring} & \rotatebox{90}{lake} & \rotatebox{90}{puddle} & \rotatebox{90}{rapids} & \rotatebox{90}{reservoir} & \rotatebox{90}{river} & \rotatebox{90}{river delta} & \rotatebox{90}{sea} & \rotatebox{90}{snow}& \rotatebox{90}{swimming pool} &  \rotatebox{90}{waterfall} & \rotatebox{90}{wetland} & \rotatebox{90}{{\bf A-acc} (\%)} & \rotatebox{90}{\bf A-mIoU} & \rotatebox{90}{{\bf acc} (\%)} & \rotatebox{90}{\bf mIoU} \\ 
			\midrule
			PSPNet & 53.8 & 29.0 & 42.9 & \underline{46.5} & 57.2 & 53.9 & 29.7 & \textbf{54.7} & 38.2 & 29.8 & 28.8 & 65.5 & \underline{63.5} & 49.9 & 47.7 & 48.4 & 47.5 & 66.19  & 46.29 & 72.72  & 40.85\\
			DeepLabv3 & 52.5 & 27.2 & \underline{52.3} & 43.8 & 58.7 & 42.5 & 31.1 & 54.2 & 46.0 & 32.4 & 27.1 & 51.1 & 61.5 & 46.3 & \textbf{53.6} & 52.8 & 52.9 & 65.83 & 46.25 & 69.21 & 36.23 \\
			DANet  & 50.5 & \textbf{34.1} & 37.1 & 37.0 & 51.0 & \underline{61.6} & 23.8 & 51.5 & 42.8 & 30.2 & 31.5 & 63.5 & 60.4 & \underline{50.8} & 43.1 & 55.2 & 54.6 & 62.00 & 45.80 & \underline{74.12} & 39.60 \\
			CCNet & 41.1 & 17.4 & 35.2 & 26.9 & 43.7 & 47.9 & 18.6 & 43.8 & 29.9 & 16.6 & 23.7 & 48.3 & 53.3 & 47.6 & 38.4 & 51.1 & 34.1 & 51.98 & 36.33 & 70.84 & 36.11 \\ 
			EMANet & 46.1 & 16.6 & 27.1 & 23.0 & 53.8 & \textbf{63.7} & 17.2 & 43.6 & 42.2 & 17.2 & 21.0 & \textbf{68.6} & 53.5 & 47.3 & 36.1 & 52.1 & 36.2 & 55.88 & 39.13 & 71.93 & 36.43\\
			ANNet & 50.9 & 22.8 & 31.6 & 32.0 & 53.1 & 58.1 & 25.6 & 52.9 & \textbf{48.4} & 20.8 & 28.6 & 56.8 & 60.4 & \textbf{51.1} & 43.9 & \textbf{57.9} & 51.4 & 61.51 & 43.90 & 74.06 & 39.79\\
			GCNet & \textbf{56.6} & 19.0 & 44.7 & 34.8 & 46.9 & 36.1 & \textbf{35.8} & 39.4 & 39.9 & \textbf{41.6} & 32.4 & 67.0 & 62.2 & 46.4 & 42.9 & 50.7 & \underline{59.7} & \textbf{69.89} & 44.48 & 68.64 & 37.73 \\
	        DNLNet & 54.4 & 26.3 & 48.8 & 36.3 & \textbf{63.2} & 55.3 & \underline{35.5} & 52.3 & 40.4 & 32.1 & 31.3 & 37.1 & 61.7 & 48.3 & \underline{52.4} & 48.7 & 54.6 & 67.72 & 45.80 & 71.95 & 39.97\\
			OCNet & \underline{56.4} & \underline{33.6} & 48.0 & 37.3 & 57.7 & 55.2 & 29.2 & 50.6 & 43.8 & 35.1 & \textbf{35.6} & \underline{65.9} & 62.7 & 47.2 & 47.9 & 53.1 & 54.9 & 67.97 & \underline{47.89} & 73.54 & \underline{41.19} \\
	        OCRNet & 52.4 & 19.4 & 46.9 & 34.9 & 48.3 & 58.8 & 30.4 & 39.7 & 42.5 & 29.8 & 31.9 & 55.5 & 55.4 & 47.3 & 43.6 & \underline{56.8} & 51.5  & 65.90 & 43.83 & 71.66  & 36.17\\ 
			\midrule
			Ours & 55.0 & 27.7 & \textbf{53.4} & \textbf{47.0} & \underline{63.1} & 60.5 & 33.2 & \underline{54.4} & \underline{46.3} & \underline{39.0} & \underline{34.7} & 63.2 & \textbf{64.2} & 50.3 & 44.9 & 53.0 & \textbf{66.1} & \underline{68.63} & \textbf{50.34} & \textbf{75.18}  & \textbf{42.22}\\
			\bottomrule
	\end{tabular}}
\end{table*}

\begin{figure*}[h]
    \makebox[\linewidth][c]{%
    \begin{subfigure}[b]{0.139\linewidth}
        \includegraphics[width=\textwidth]{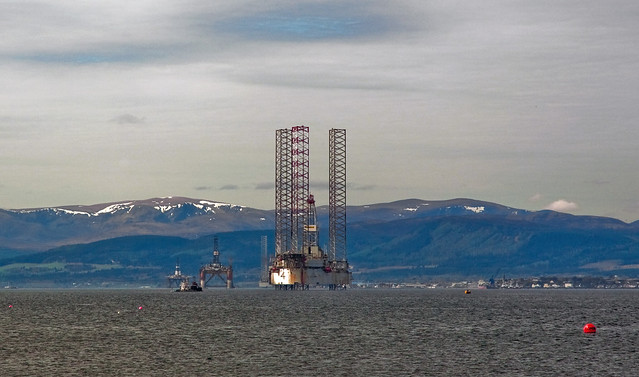}
    \end{subfigure}
    \begin{subfigure}[b]{0.139\linewidth}
        \includegraphics[width=\textwidth]{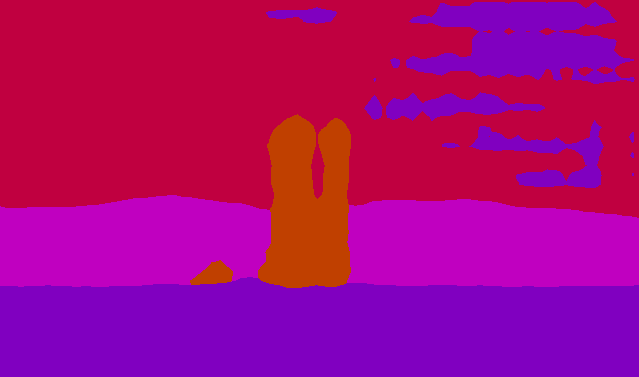}
    \end{subfigure}
    \begin{subfigure}[b]{0.139\linewidth}
        \includegraphics[width=\textwidth]{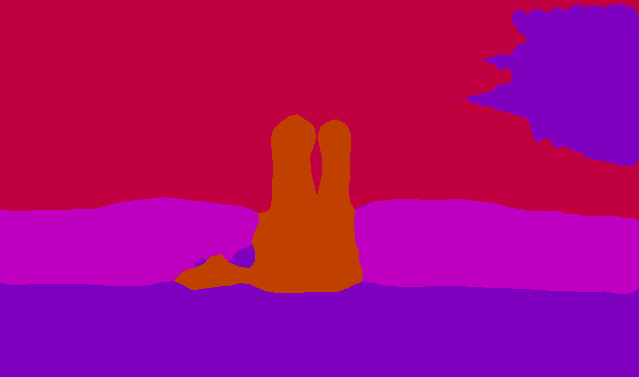}
    \end{subfigure}
    \begin{subfigure}[b]{0.139\linewidth}
        \includegraphics[width=\textwidth]{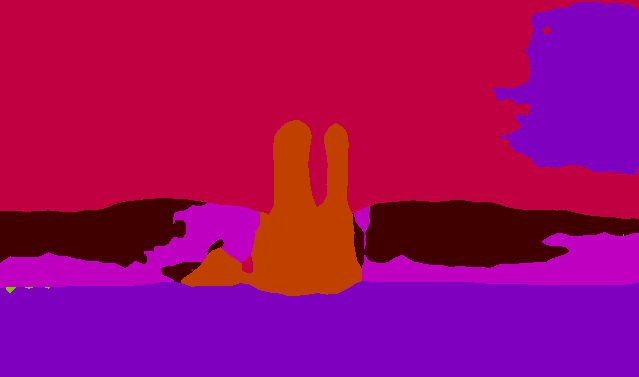}
    \end{subfigure}
    \begin{subfigure}[b]{0.139\linewidth}
        \includegraphics[width=\textwidth]{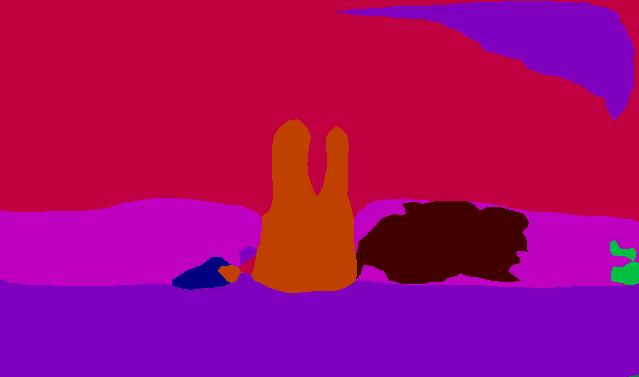}
    \end{subfigure}
    \begin{subfigure}[b]{0.139\linewidth}
        \includegraphics[width=\textwidth]{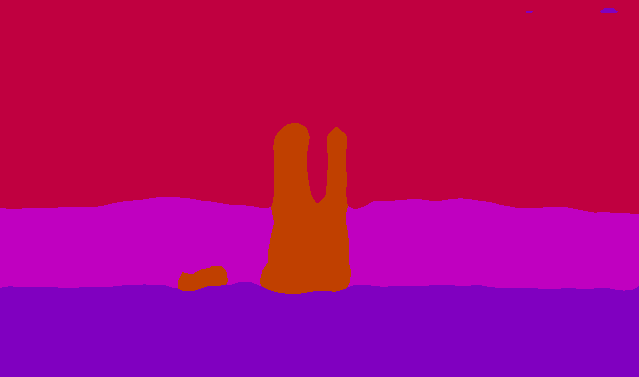}
    \end{subfigure}
    \begin{subfigure}[b]{0.139\linewidth}
        \includegraphics[width=\textwidth]{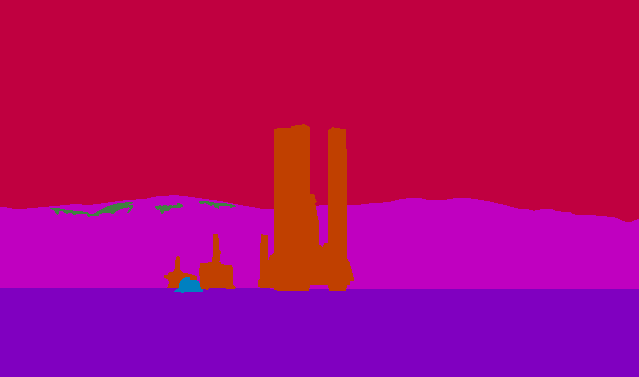}
    \end{subfigure}}
    
    \makebox[\linewidth][c]{%
    \begin{subfigure}[b]{0.139\linewidth}
        \includegraphics[width=\textwidth]{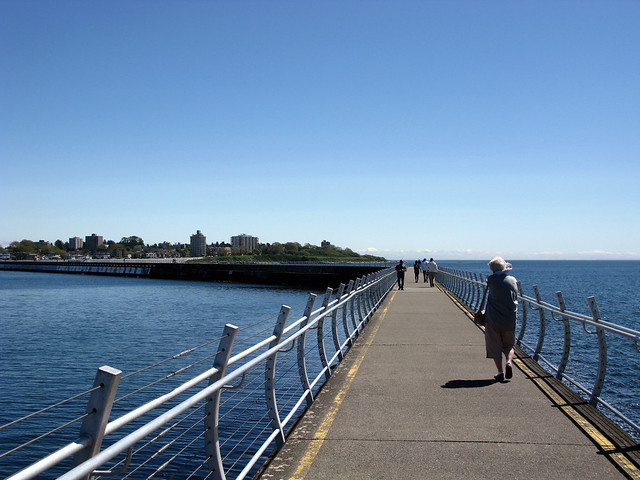}
    \end{subfigure}
    \begin{subfigure}[b]{0.139\linewidth}
        \includegraphics[width=\textwidth]{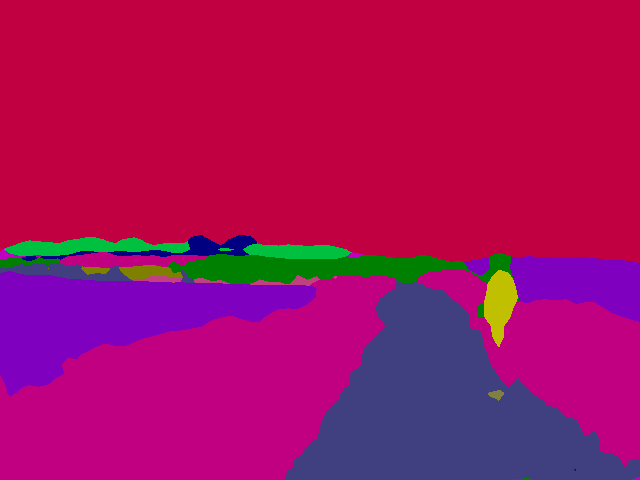}
    \end{subfigure}
    \begin{subfigure}[b]{0.139\linewidth}
        \includegraphics[width=\textwidth]{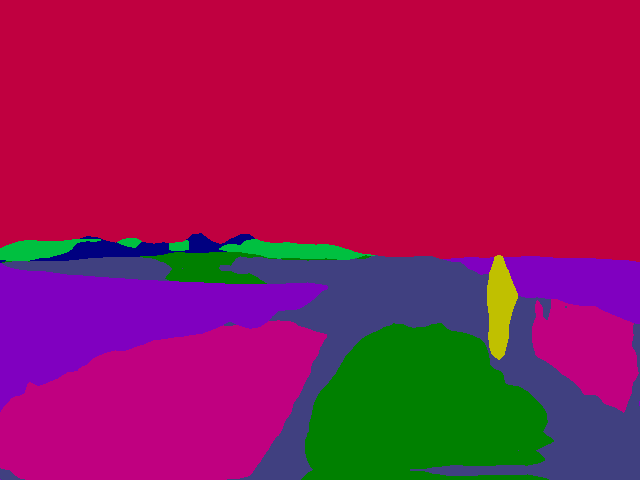}
    \end{subfigure}
    \begin{subfigure}[b]{0.139\linewidth}
        \includegraphics[width=\textwidth]{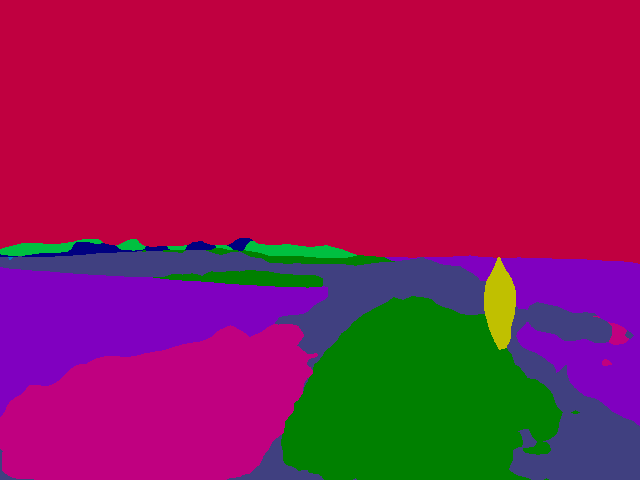}
    \end{subfigure}
    \begin{subfigure}[b]{0.139\linewidth}
        \includegraphics[width=\textwidth]{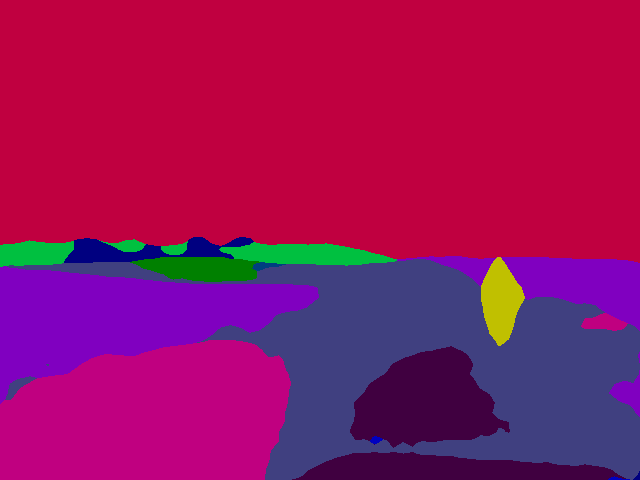}
    \end{subfigure}
    \begin{subfigure}[b]{0.139\linewidth}
        \includegraphics[width=\textwidth]{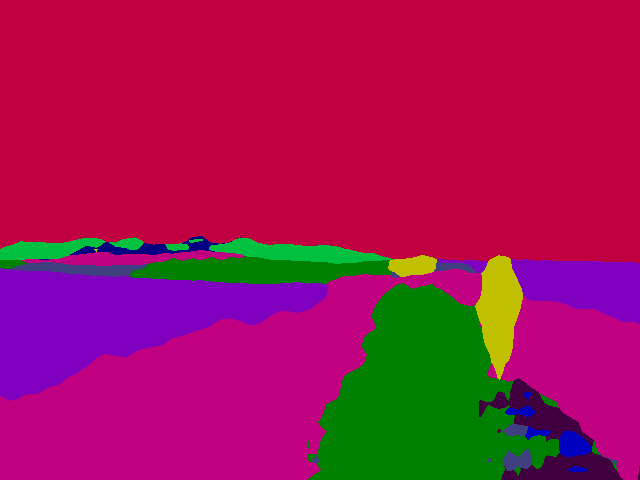}
    \end{subfigure}
    \begin{subfigure}[b]{0.139\linewidth}
        \includegraphics[width=\textwidth]{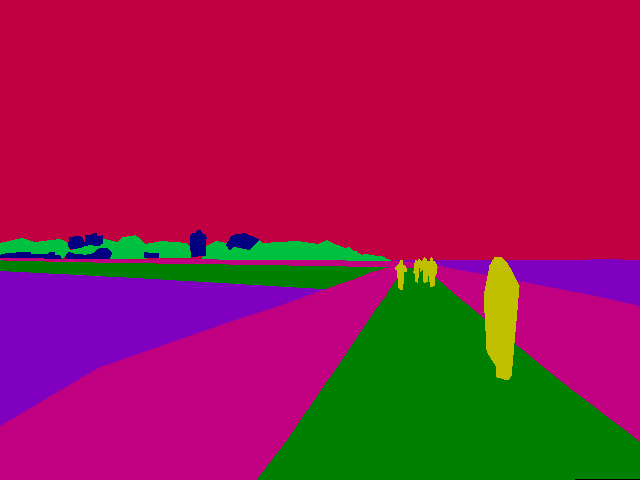}
    \end{subfigure}}
    
    \makebox[\linewidth][c]{%
    \begin{subfigure}[b]{0.139\linewidth}
        \includegraphics[width=\textwidth]{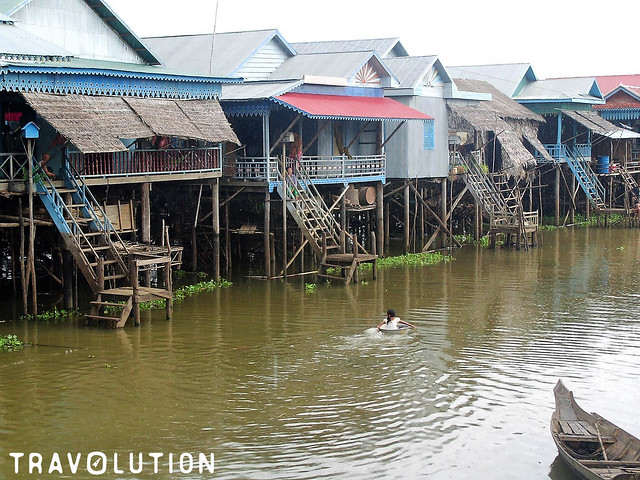}
    \end{subfigure}
    \begin{subfigure}[b]{0.139\linewidth}
        \includegraphics[width=\textwidth]{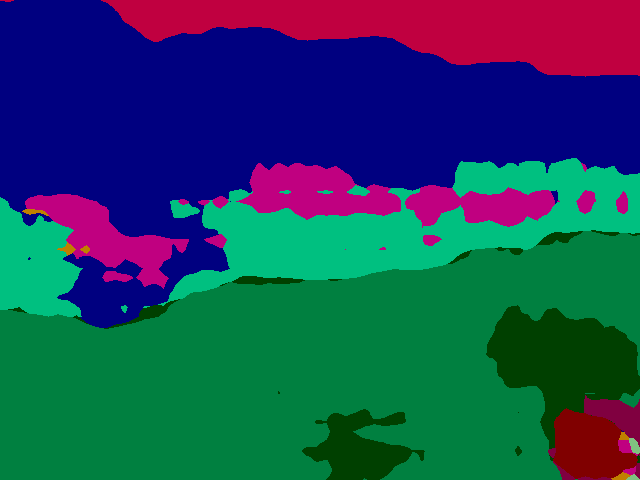}
    \end{subfigure}
    \begin{subfigure}[b]{0.139\linewidth}
        \includegraphics[width=\textwidth]{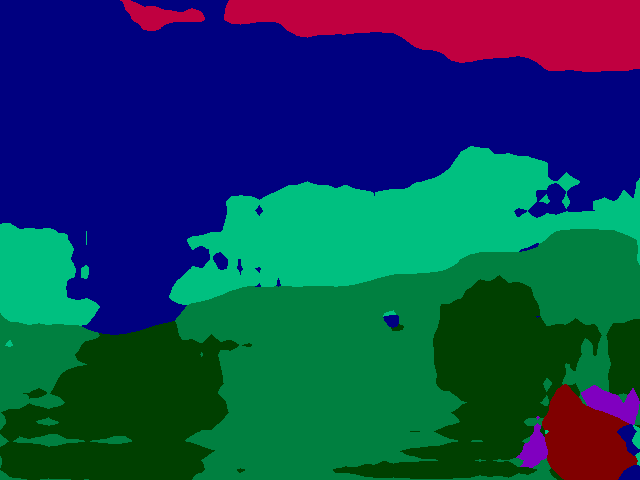}
    \end{subfigure}
    \begin{subfigure}[b]{0.139\linewidth}
        \includegraphics[width=\textwidth]{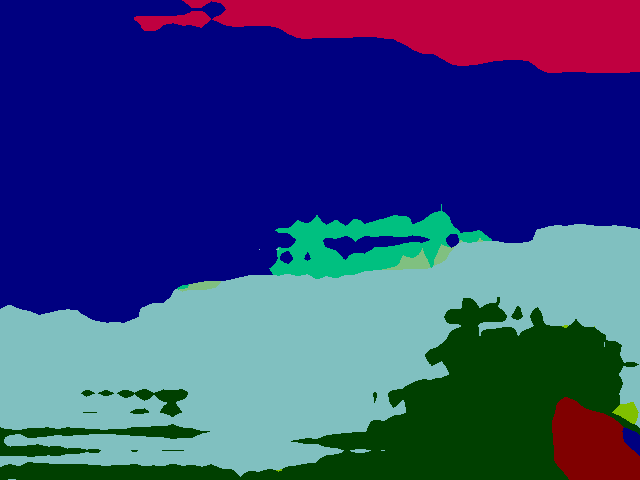}
    \end{subfigure}
    \begin{subfigure}[b]{0.139\linewidth}
        \includegraphics[width=\textwidth]{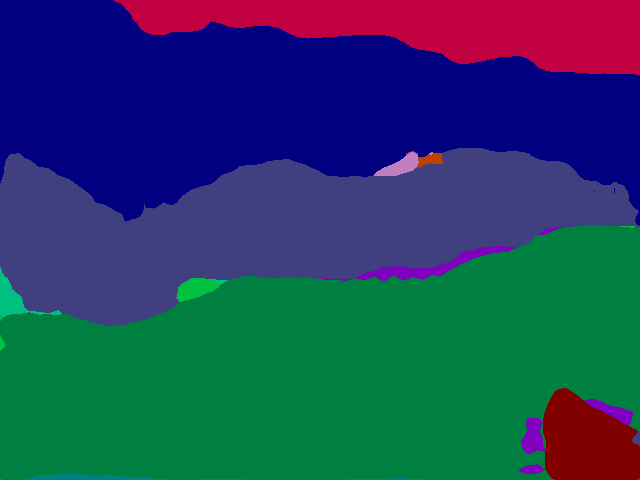}
    \end{subfigure}
    \begin{subfigure}[b]{0.139\linewidth}
        \includegraphics[width=\textwidth]{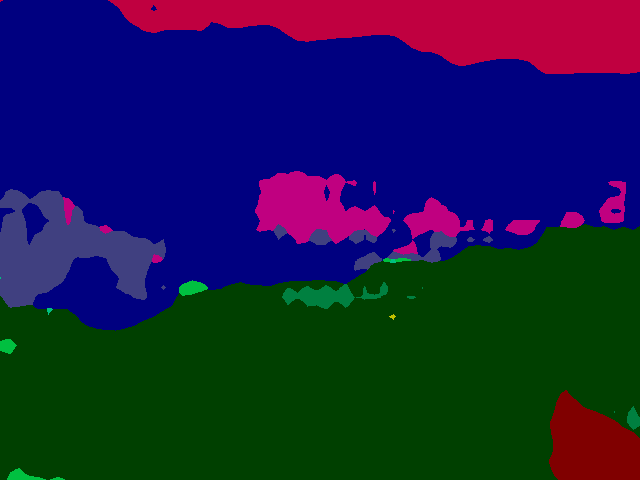}
    \end{subfigure}
    \begin{subfigure}[b]{0.139\linewidth}
        \includegraphics[width=\textwidth]{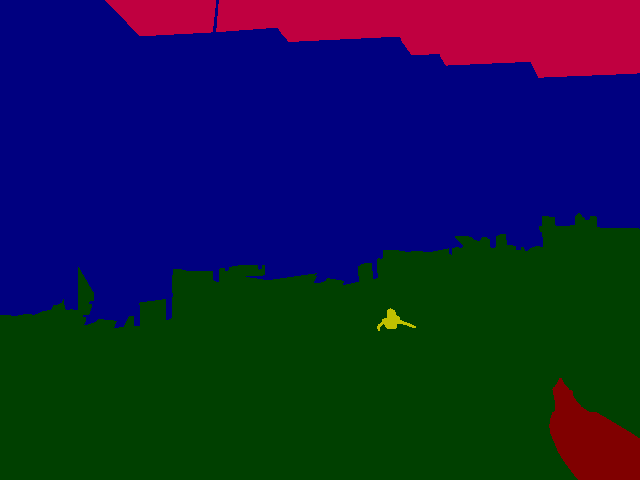}
    \end{subfigure}}
    
    \makebox[\linewidth][c]{%
    \begin{subfigure}[b]{0.139\linewidth}
        \includegraphics[width=\textwidth]{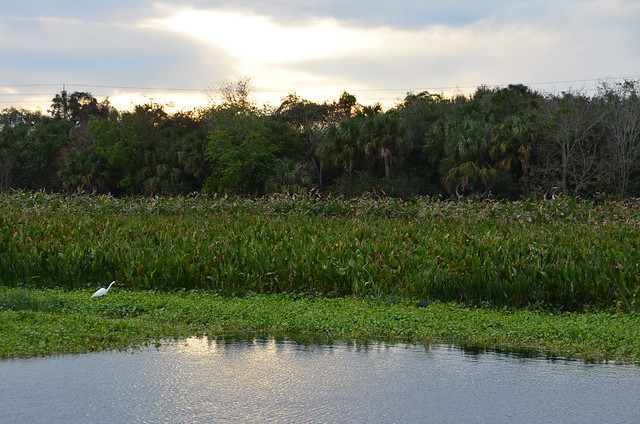}
        \vspace{-15pt} \caption*{Image}
    \end{subfigure}
    \begin{subfigure}[b]{0.139\linewidth}
        \includegraphics[width=\textwidth]{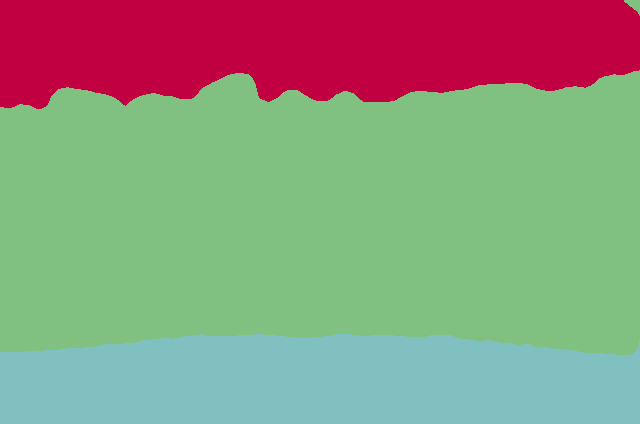}
        \vspace{-15pt} \caption*{OCNet}
    \end{subfigure}
    \begin{subfigure}[b]{0.139\linewidth}
        \includegraphics[width=\textwidth]{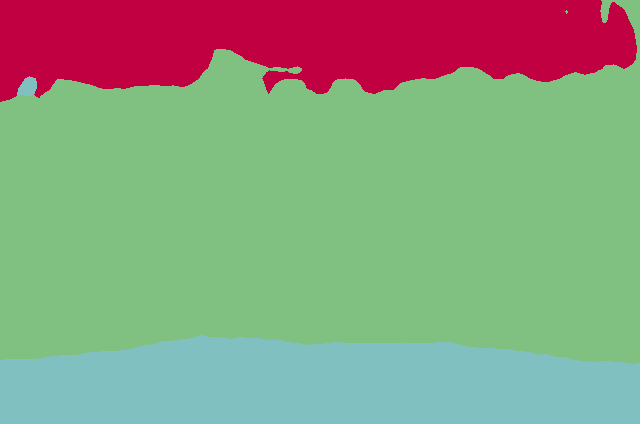}
        \vspace{-15pt} \caption*{GCNet}
    \end{subfigure}
    \begin{subfigure}[b]{0.139\linewidth}
        \includegraphics[width=\textwidth]{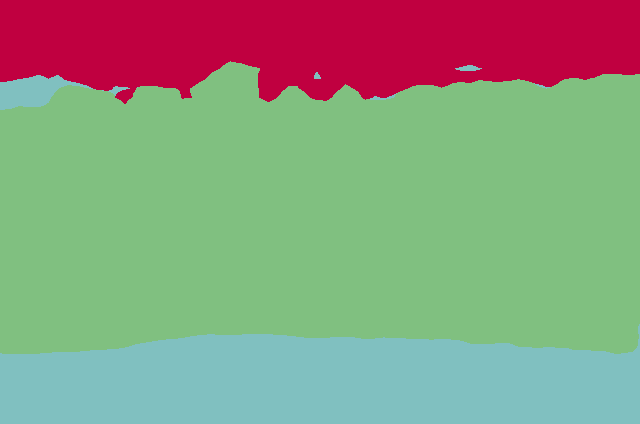}
        \vspace{-15pt} \caption*{PSPNet}
    \end{subfigure}
    \begin{subfigure}[b]{0.139\linewidth}
        \includegraphics[width=\textwidth]{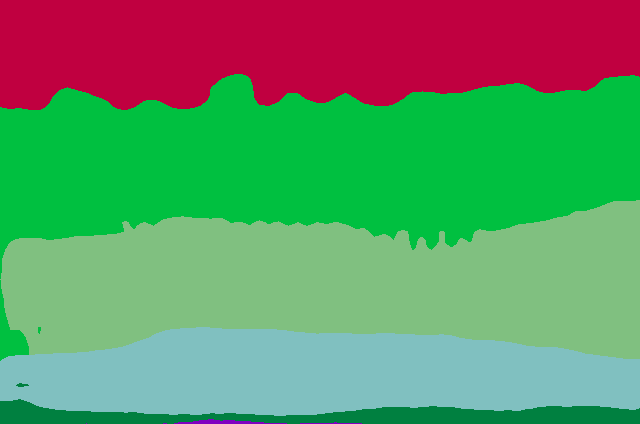}
        \vspace{-15pt} \caption*{DANet}
    \end{subfigure}
    \begin{subfigure}[b]{0.139\linewidth}
        \includegraphics[width=\textwidth]{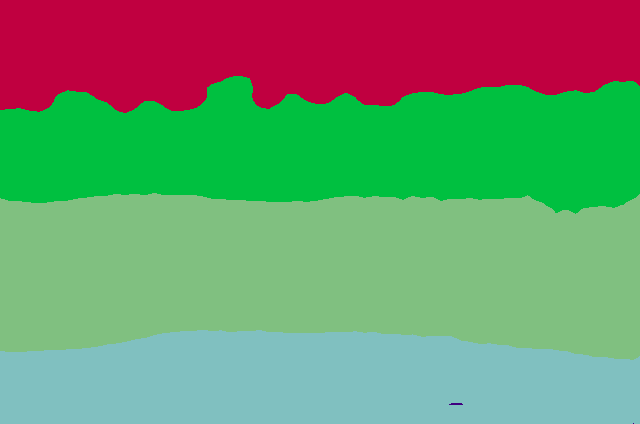}
        \vspace{-15pt} \caption*{AQUANet}
    \end{subfigure}
    \begin{subfigure}[b]{0.139\linewidth}
        \includegraphics[width=\textwidth]{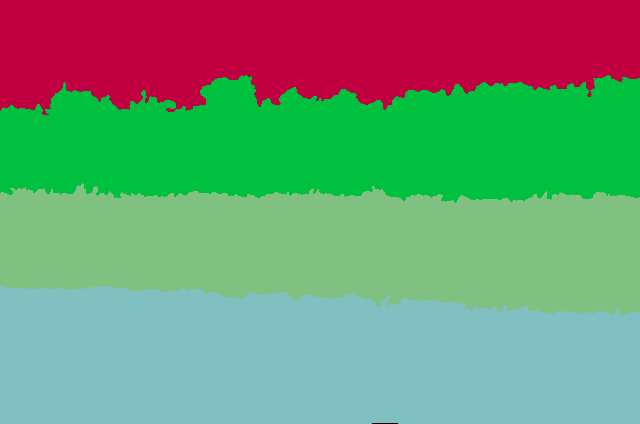}
        \vspace{-15pt} \caption*{GT}
    \end{subfigure}}
    \caption{Visualisation comparison of AQUANet and four well-known methods on the ATLANTIS validation set.}
    \label{fig:56test}
\end{figure*}

\subsection{Experimental settings}
\label{exp_setting}

The AQUANet is implemented using PyTorch. 
During training, the base learning rate is set to $2.5 \times 10^{-4}$ and it is decayed following the poly policy \cite{zhao2017pyramid}.
The network is optimized using SGD with a momentum of 0.9 and weight decay of 0.0001.
In total, we train the network for 30 epochs, around 80K iterations with a batch size of 2.
The training data are augmented with random horizontal flipping, random scaling ranging from 0.5 to 2.0 and random cropping with the size of $640 \times 640$.

\subsection{Comparisons}
We use several state-of-the-art networks to perform training and testing on ATLANTIS, including PSPNet \cite{zhao2017pyramid}, DeepLabv3 \cite{chen2017rethinking}, CCNet \cite{huang2019ccnet}, EMANet \cite{li2019expectation}, ANNet \cite{zhu2019asymmetric}, DANet \cite{fu2019dual}, DNLNet \cite{yin2020disentangled}, GCNet \cite{cao2019gcnet}, OCNet \cite{YuanW18}, OCRNet \cite{yuan2019object}. 
For a fair comparison, we train all the networks with the same backbone (ResNet-101) for 30 epochs.
As shown in Table~\ref{tab:test}, the proposed AQUANet outperforms all these networks on waterbody image 
semantic segmentation.
Figure~\ref{fig:56test} shows the visualization results of some samples from ATLANTIS validation set. Considering the ground truth and comparing with other networks' outputs, the boundaries between different classes are better preserved in AQUANet output.
Compared with \citeauthor{chen2017rethinking}~\cite{chen2017rethinking}, \citeauthor{zhao2017pyramid}~\cite{zhao2017pyramid}, \citeauthor{fu2019dual}~\cite{fu2019dual}, and \citeauthor{YuanW18}~\cite{YuanW18}, 
our method achieves better results on both the aquatic and non-aquatic regions. 

\subsection{Failure cases}
\begin{figure}[h]
    \begin{subfigure}[b]{0.328\linewidth}
        \includegraphics[width=\textwidth]{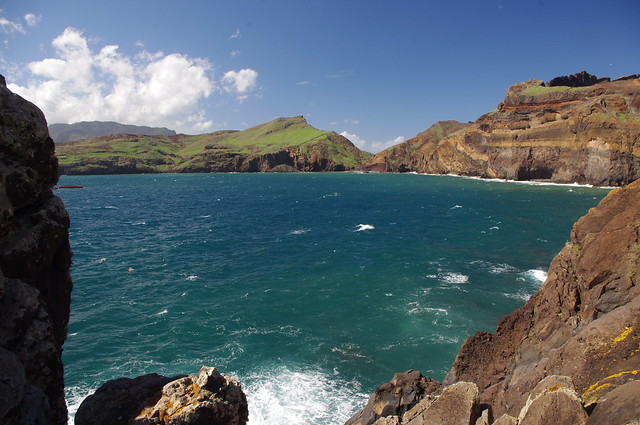}
    \end{subfigure}
    \begin{subfigure}[b]{0.328\linewidth}
        \includegraphics[width=\textwidth]{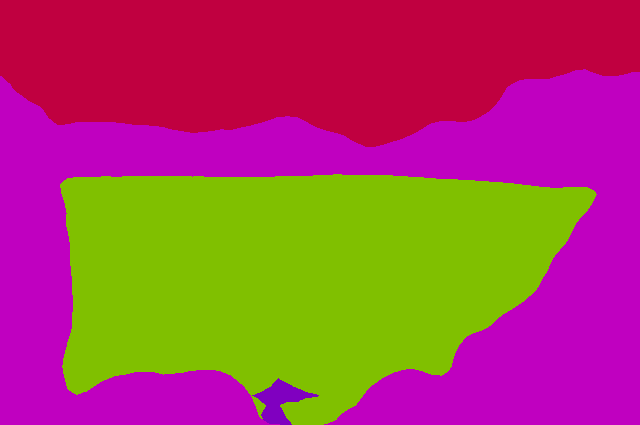}
    \end{subfigure}
    \begin{subfigure}[b]{0.328\linewidth}
        \includegraphics[width=\textwidth]{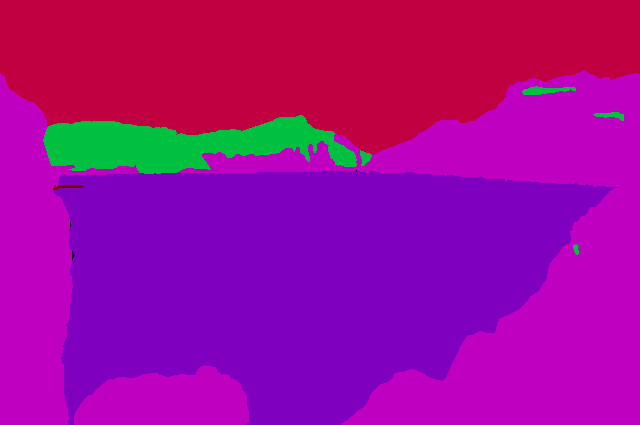}
    \end{subfigure}
    \begin{subfigure}[b]{0.328\linewidth}
        \includegraphics[width=\textwidth]{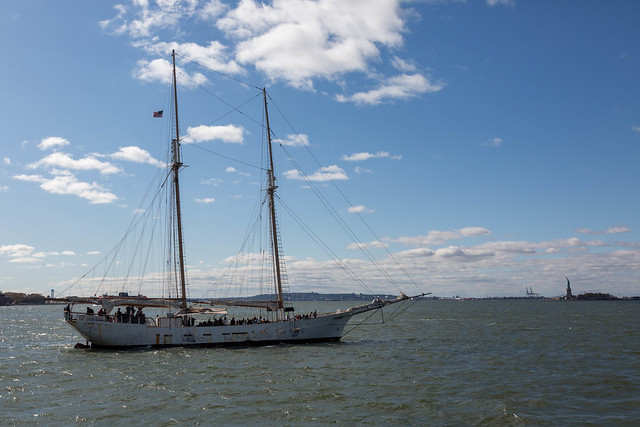}
    \end{subfigure}
    \begin{subfigure}[b]{0.328\linewidth}
        \includegraphics[width=\textwidth]{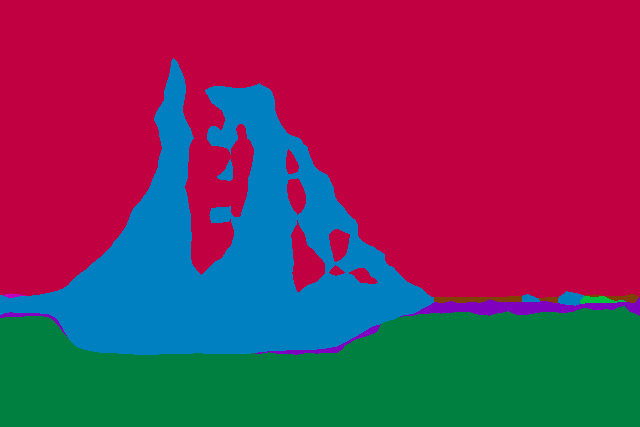}
    \end{subfigure}
    \begin{subfigure}[b]{0.328\linewidth}
        \includegraphics[width=\textwidth]{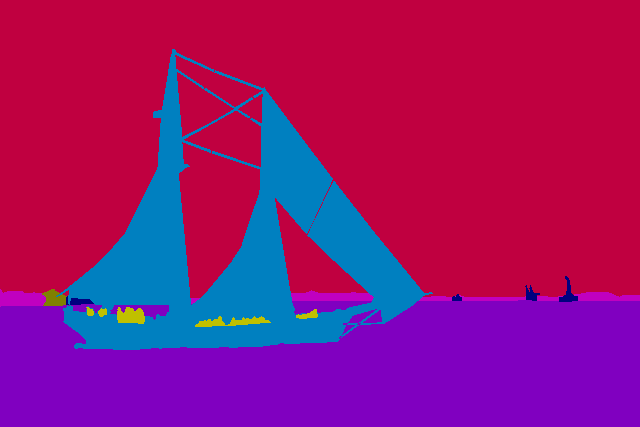}
    \end{subfigure}
    \begin{subfigure}[b]{0.328\linewidth}
        \includegraphics[width=\textwidth]{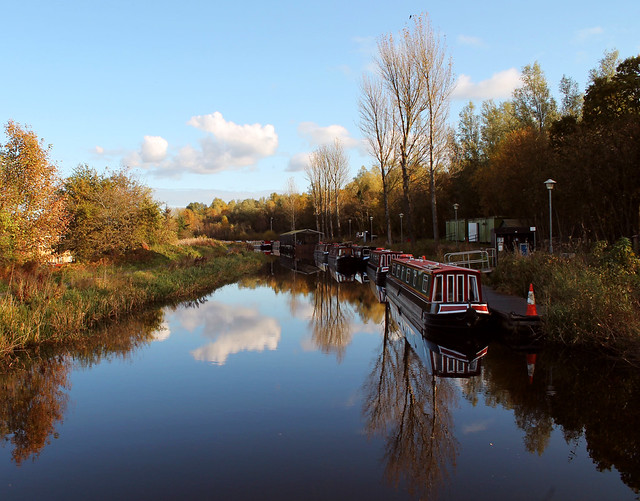}
        \vspace{-15pt} \caption*{Image}
    \end{subfigure}
    \begin{subfigure}[b]{0.328\linewidth}
        \includegraphics[width=\textwidth]{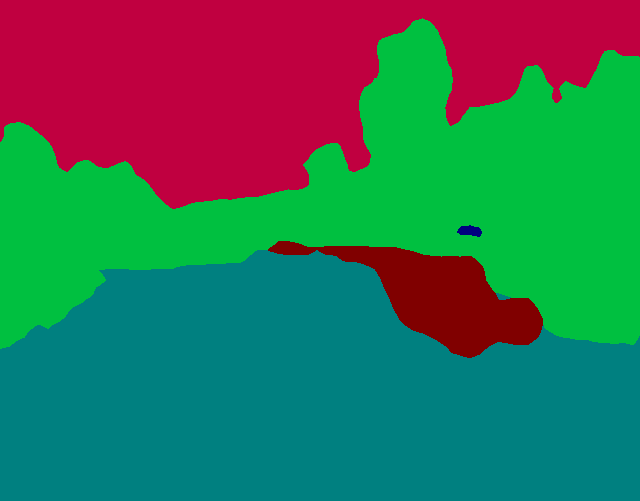}
        \vspace{-15pt} \caption*{Ours}
    \end{subfigure}
    \begin{subfigure}[b]{0.328\linewidth}
        \includegraphics[width=\textwidth]{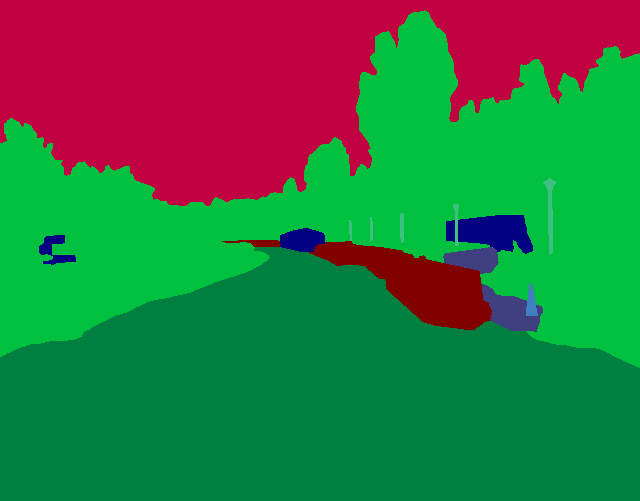}
        \vspace{-15pt} \caption*{GroundTruth}
    \end{subfigure}
    \caption{Failure cases from the ATLANTIS val set.}
    \label{fig:fail}
\end{figure}

Due to the challenges associated with segmentation of waterbody images, there are still many failure cases we found in the testing stage.
Three failure examples are shown Figure~\ref{fig:fail}, from which we observe that many aquatic classes are vulnerable to mis-classification -- here sea is mis-classified to lake and river (row 1-2) and river is mis-classified to canal (row 3).

\subsection{Ablation studies}
We also conduct ablation studies to compare a number of different model variants of the proposed network, including the design of aquatic and non-aquatic paths, and the two feature modulations. The results are shown in Table~\ref{tab:ablation}. We can see that the design of two paths can improve the performance of waterbody image semantic segmentation. Moreover, both the proposed low-level feature modulation (LM) and the cross-path modulation (CM) can achieve certain performance gains in terms of acc and mIoU.

\begin{table}[h]
	\centering
	\caption{Ablation study of each proposed component of AQUANet on the ATLANTIS dataset.}
    \renewcommand\arraystretch{1.0}
		{\begin{tabular}{ccc|cccc}
			\toprule
			Two Paths & LM & CM &A-acc &A-mIoU & acc & mIoU  \\
			\midrule
			 & & & 67.27 & 44.53 &73.28 &38.81 \\
			$\surd$ & & & 67.73 & 47.89 & 75.29 &40.28\\
			$\surd$ &$\surd$ & & 68.11 & 47.90 &75.29 &40.28\\
			$\surd$ & &$\surd$ & 67.21 &46.63 &75.81 &40.57  \\
			$\surd$ &$\surd$ &$\surd$ & 68.85 & 48.42 & 76.18 &40.83 \\
			\bottomrule

	\end{tabular}}
	\label{tab:ablation}
\end{table}

Comparing to other state-of-the-art semantic segmentation models, AQUANet provides highest mIoU and accuracy. In addition, by considering just labels which includes water content, AQUANet still works better than others (mIoU=50.34$\%$). In this regard, OCNet provides the second highest performance (mIoU=47.89$\%$) and GCNet also provides the best results for canal, lake and reservoir. These results approved AQUANet is well customized for analysis of waterbodis and water-related scenes. However, considering mIoU as a more informative metric for semantic segmentation task, all approaches could only achieve 36.11\%$\sim$42.22\% mIoU on the proposed dataset. It shows that the semantic segmentation of waterbodies and related objects is a challenging task and needs more researches.

\subsection{ATeX Experimental results}
We further train ten well-known classification models including VGG~\cite{simonyan2014very}, ResNet~\cite{he2016deep}, SqueezeNet~\cite{iandola2016squeezenet}, DenseNet~\cite{huang2017densely}, GoogLeNet~\cite{szegedy2015going}, ShuffleNet v2~\cite{ma2018shufflenet}, MobileNetV2~\cite{sandler2018mobilenetv2}, ResNeXt~\cite{xie2017aggregated}, Wide ResNet~\cite{zagoruyko2016wide} and EfficientNet~\cite{tan2019efficientnet} on the proposed ATeX dataset. All models are implemented using PyTorch. Cross-entropy loss function is applied for training networks. We train all the networks with the same 30 training epochs, SGD optimizer with a momentum of 0.9 and weight decay of 0.0001, and batch size is set to 64. For all networks, the learning rate is first set to $2.5 \times 10^{-4}$, then it is adjusted based on decaying rate of the resulted loss function during training. Table~\ref{tab:model_results} shows the training time and learning rate for each models over certain 30 epochs.

\begin{table}
    \centering
    \small
    \caption{The perfomance result on ATeX test set by well-known classification models.}
    \label{tab:model_results}
    \begin{tabular}{lcccccc}
        \toprule
        \multirow{2}{1pt}{Networks} & Time & \multirow{2}{*}{LR} & Val & \multicolumn{3}{c}{Test} \\
            & [mm:ss] &   & Acc. & Prec. & Recall & F1 \\
        \midrule
        Wide-ResNet-50-2 & 06:56 & 2.5E-4 & 91 & 77 & 75 & 75\\
        VGG-16 & 04:38 & 2.5E-4 & 90 & 75 & 72 & 72\\
        SqueezeNet & 00:47 & 7.5E-4 & 82 & 81 & 81 & 81\\
        ShuffleNet V2 $\times 1.0$ & 01:46 & 1.0E-2 & 90 & 90 & 90 & 90\\
        ResNeXt-50-$32 \times 4d$ & 03:15 & 2.5E-4 & 90 & 77 & 75 & 75\\
        ResNet-18 & 01:28 & 2.5E-4 & 87 & 74 & 72 & 72\\
        MobileNet V2 & 01:35 & 2.5E-4 & 88 & 74 & 72 & 72\\
        GoogLeNet & 01:51 & 5.0E-3 & 89 & 88 & 88 & 88\\
        EffNet-B7 & 12:42 & 1.0E-2 & 90 & 91 & 91 & 91\\
        EffNet-B0 & 02:38 & 7.5E-3 & 91 & 90 & 90 & 90\\
        DenseNet-161 & 06:15 & 2.5E-4 & 91 & 81 & 79 & 79\\
        \bottomrule
    \end{tabular}
\end{table}

Three common performance metrics including Precision, Recall and F1-score are reported to evaluate the performance of the models on ATeX. Table~\ref{tab:model_results} shows weighted average (averaging the support-weighted mean per label) of these three metrics on the test set. Accordingly, EffNet-B7, EffNet-B0 and ShuffleNet V2 $\times 1.0$ provide the best results. Considering training time, ShuffleNet V2 $\times 1.0$ can be presented as the most efficient network.

\section{Conclusion} \label{sec:conclusion}
In this paper, we introduced ATLANTIS, a large-scale dataset for semantic segmentation of waterbodies and water-related scenes, by carefully collecting images of diverse area from the internet and providing high-quality annotations with the help of annotators major in water resources engineering. We further provided comprehensive analysis of the characteristic of ATLANTIS and reported the performance of current state-of-the-arts by training and testing the networks on our dataset. A novel baseline network AQUANet is also proposed for waterbody image semantic segmentation and achieves the best performance on ATLANTIS. Additionally, we constructed ATLANTIS Texture (ATeX) dataset that are sampled from ATLANTIS for texture-based classification and evaluated several baseline methods on it.

ATLANTIS posed significant challenges for semantic segmentation which we believe will boost new insights in both water resources engineering and computer vision communities.


\bibliographystyle{plainnat}  
\bibliography{refs}

\end{document}